\documentclass[letterpaper, 10 pt, conference]{ieeeconf}

\IEEEoverridecommandlockouts
\usepackage{amsmath, amssymb, amsfonts}

\usepackage{graphicx}
\usepackage{booktabs}
\usepackage{multirow}
\usepackage{array}

\usepackage[dvipsnames]{xcolor}
\usepackage[colorlinks=true, linkcolor=MidnightBlue, citecolor=OliveGreen, urlcolor=BrickRed]{hyperref}

\usepackage{cite}

\let\labelindent\relax  
\usepackage{enumitem}
\usepackage{url}
\usepackage{algorithm}
\usepackage{algorithmic}
\usepackage{tikz}
\usetikzlibrary{arrows.meta, positioning, shapes.geometric, calc, fit, shadows, backgrounds}
\usepackage{pgfplots}
\pgfplotsset{compat=1.18}
\usepgfplotslibrary{fillbetween}

\newcommand{\Lref}{L_{\mathrm{ref}}}
\newcommand{\Pmode}{P_{\mathrm{mode}}}
\newcommand{\GSDpred}{\mathrm{GSD}_{\mathrm{pred}}}
\newcommand{\GSDgt}{\mathrm{GSD}_{\mathrm{gt}}}
\newcommand{\GSDmax}{\mathrm{GSD}_{\max}}
\newcommand{\Pthresh}{P_{\mathrm{thresh}}}
\newcommand{\GPTfo}{GPT-4{\sffamily o}}
\newcommand{\YOLOviiil}{YOLOv8\textit{l}-OBB}
\newcommand{\YOLOxil}{YOLO11\textit{l}-OBB}

\title{\LARGE \bf VehAnchor: Metadata-Free Metric Scale Recovery\\from Vehicle Cues in Aerial Imagery}

\author{\parbox{\textwidth}{\centering
Yifei Chen$^{1}$, Chenqian Le$^{2}$, Jiayi Cheng$^{2}$, and Xupeng Chen$^{1,*}$}%
\thanks{$^{1}$Yifei Chen and Xupeng Chen are with Dimension Gate,
        Haidian District, Beijing 100083, China.}%
\thanks{$^{2}$Chenqian Le and Jiayi Cheng are with New York University,
        New York, NY, USA.}%
\thanks{$^{*}$Corresponding author: Xupeng Chen.}%
}

\begin{document}
\maketitle
\thispagestyle{empty}
\pagestyle{empty}

\begin{abstract}
Autonomous aerial robots operating in GPS-denied or communication-degraded environments frequently lose access to camera metadata and telemetry, leaving onboard perception systems unable to recover the absolute metric scale of the scene.
As LLM/VLM-based planners are increasingly adopted as high-level agents for embodied systems, their ability to reason about physical dimensions becomes safety-critical---yet our experiments show that five state-of-the-art VLMs suffer from \emph{spatial scale hallucinations}, with median area estimation errors exceeding 50\%.
We propose VehAnchor, a lightweight, deterministic \emph{Geometric Perception Skill} designed as a callable tool that any LLM-based agent can invoke to recover Ground Sample Distance (GSD) from ubiquitous environmental anchors: small vehicles detected via oriented bounding boxes, whose modal pixel length is robustly estimated through kernel density estimation and converted to GSD using a pre-calibrated reference length.
The tool returns both a GSD estimate and a composite confidence score, enabling the calling agent to autonomously decide whether to trust the measurement or fall back to alternative strategies.
On the DOTA~v1.5 benchmark, VehAnchor achieves 6.87\% median GSD error on 306~images.
Integrated with SAM-based segmentation for downstream area measurement, the pipeline yields 19.7\% median error on a 100-entry benchmark---with 2.6$\times$ lower category dependence and 4$\times$ fewer catastrophic failures than the best VLM baseline---demonstrating that equipping agents with deterministic geometric tools is essential for safe autonomous spatial reasoning.
\end{abstract}

\begin{figure*}[t]
  \centering
  \includegraphics[width=\textwidth]{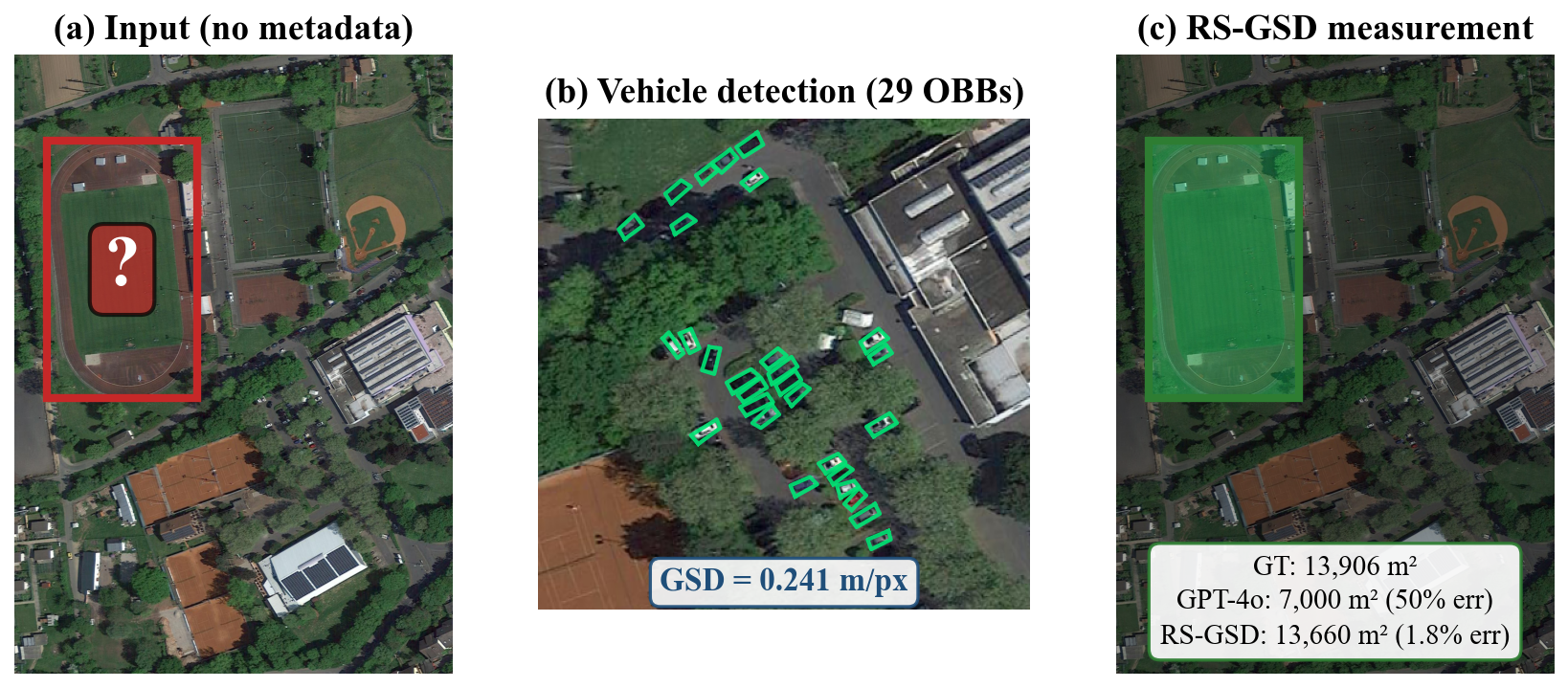}
  \caption{Our Geometric Perception Skill in action on a DOTA~v1.5 image (P2593, Ground Track Field). (a)~A UAV captures an image without metadata; the target object (red box) must be measured from pixels alone. (b)~Our skill detects small vehicles via oriented bounding boxes (green) and derives $\mathrm{GSD} = 0.241$\,m/px from their modal pixel length. (c)~Combined with SAM segmentation, the pipeline predicts the field area within 1.8\% of ground truth, whereas \GPTfo{} hallucinates a 50\% underestimate.}
  \label{fig:teaser}
\end{figure*}

\section{INTRODUCTION}
\label{sec:intro}

Micro aerial vehicles (MAVs) and unmanned aerial vehicles (UAVs) are increasingly deployed for autonomous tasks---disaster assessment, infrastructure inspection, precision agriculture, and search-and-rescue---where onboard perception must deliver metric-scale spatial understanding in real time.
In GPS-denied, communication-degraded, or previously unmapped environments, these platforms routinely lose access to camera metadata and absolute telemetry, leaving monocular imagery as the sole input.
Without knowing the Ground Sample Distance (GSD)---the physical size of each pixel---pixel-level measurements cannot be converted to real-world dimensions, and any downstream spatial reasoning becomes unreliable.

The robotics community has increasingly turned to vision-language models (VLMs) and large language models (LLMs) as high-level planners and agents for embodied systems~\cite{chen2024spatialvlm}.
However, our experiments reveal a critical failure mode: when tasked with estimating physical areas from aerial imagery alone, five state-of-the-art VLMs exhibit median errors of 38--52\%, with frequent order-of-magnitude deviations.
We term this failure \emph{Spatial Scale Hallucination}---the systematic inability of VLMs to ground visual observations in metric scale without explicit geometric calibration.
This poses a direct safety risk for autonomous UAV operations: a planner that misjudges a landing zone's dimensions by 50\% could attempt a catastrophic landing on an insufficiently sized surface.

We propose VehAnchor---a lightweight, deterministic \emph{Geometric Perception Skill} and callable tool that LLM-based planners can invoke to recover absolute scale from monocular aerial imagery.
Our key observation is that small vehicles are among the most ubiquitous objects in urban and suburban scenes, with physical lengths concentrated around 4--5\,m worldwide.
The pipeline detects vehicles via oriented bounding boxes (OBB), robustly estimates their modal pixel length through kernel density estimation (KDE), and converts it to GSD using a pre-calibrated reference length.
Figure~\ref{fig:teaser} previews the approach: from a single metadata-free image, VehAnchor anchors scale on detected vehicles and, combined with segmentation, measures a target region within 1.8\% of ground truth where \GPTfo{} underestimates by 50\%.

Our contributions are:
\begin{itemize}[nosep]
  \item A metadata-free GSD estimation method using vehicles as geometric anchors, achieving 6.87\% median error on DOTA~v1.5 (306~images, 67\% coverage), with KDE providing 17\% improvement over mean aggregation.
  \item Empirical evidence of \emph{Spatial Scale Hallucination} in VLMs: a 100-entry benchmark showing that even with explicit vehicle-length hints, VLMs exhibit 2.6$\times$ higher category dependence and 4$\times$ more catastrophic failures than our deterministic pipeline.
  \item Integration as a \emph{Geometric Perception Skill} for tool-augmented agents: a stateless API that accepts a monocular image and returns ($\GSDpred$, $\mathcal{C}$), enabling LLM/VLM planners to make safety-aware metric decisions without GPS or metadata dependencies.
\end{itemize}

\section{RELATED WORK}
\label{sec:related}

\paragraph{GSD and scale estimation.}
Traditional GSD determination relies on sensor specifications, flight altitude, and camera calibration \cite{wolf2014elements, toutin2004geometric}.
When metadata is unavailable, alternative approaches attempt to infer spatial scale from image content: shadow-based methods~\cite{liasis2016shadow}, monocular depth estimation~\cite{ranftl2022midas, bhat2023zoedepth}, and regression CNNs~\cite{lee2019gsd}.
However, these methods either require supervised training with GSD labels or recover only relative depth.
Wang and Wang~\cite{wang2025visual} proposed a visual agentic system for scale estimation using \GPTfo{}-based reference object selection, but their dependence on a proprietary model limits reproducibility.
We address these gaps by focusing on a single, well-characterised reference class (small vehicles) with kernel density estimation~\cite{rosenblatt1956remarks, parzen1962estimation, silverman1986density} for robust mode finding.

\paragraph{VLM spatial reasoning deficiencies.}
Despite rapid progress in vision-language models for remote sensing~\cite{kuckreja2024geochat, li2024visionlanguage, liu2023remoteclip}, fundamental limitations in spatial reasoning persist.
Liu et al.~\cite{liu2023vsr} demonstrate a large human--model gap on spatial relation tasks; Chen et al.~\cite{chen2024spatialvlm} show VLMs lack 3D spatial reasoning for distance and size estimation; and Liao et al.~\cite{liao2024qspatial} find that VLMs struggle with quantitative spatial reasoning even when provided with reference objects.
These findings motivate our approach: rather than relying on VLMs for metric-scale reasoning, we provide a deterministic geometric tool that embodied agents can invoke to ground their spatial understanding.

\section{METHOD}
\label{sec:method}

\subsection{Overview}

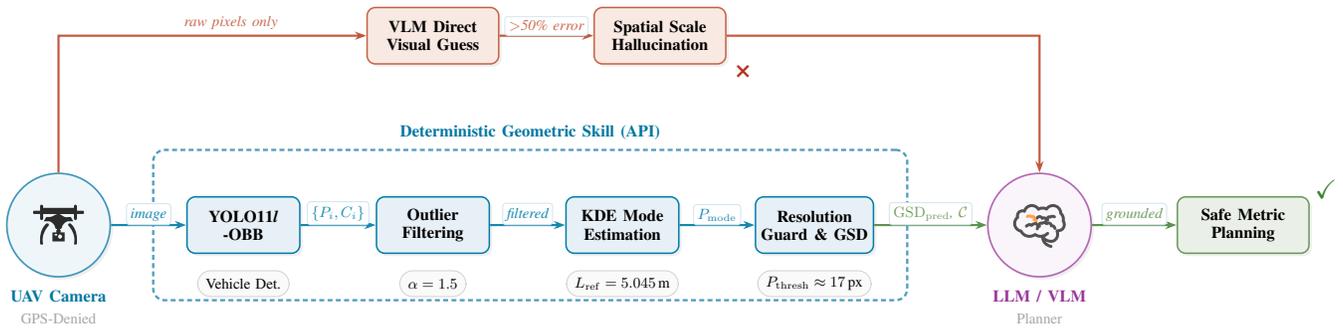
\begin{figure*}[t]
  \centering
  \resizebox{\textwidth}{!}{%
  \begin{tikzpicture}[
    box/.style={
      rectangle, draw=#1!80, fill=#1!6,
      minimum height=1.2cm, minimum width=2.4cm, align=center,
      font=\normalsize, rounded corners=5pt, line width=0.9pt,
      drop shadow={shadow xshift=0.6mm, shadow yshift=-0.6mm, fill=black!12}
    },
    box/.default=MidnightBlue,
    skillbox/.style={
      rectangle, draw=MidnightBlue!55, fill=white,
      align=center, font=\normalsize, rounded corners=7pt,
      line width=1.4pt, densely dashed
    },
    iconnode/.style={
      circle, draw=#1!60, fill=#1!4,
      minimum size=2.2cm, align=center, line width=1.0pt,
      drop shadow={shadow xshift=0.6mm, shadow yshift=-0.6mm, fill=black!12}
    },
    iconnode/.default=MidnightBlue,
    outcome/.style={
      rectangle, draw=#1!75, fill=#1!8,
      minimum height=1.2cm, minimum width=2.8cm, align=center,
      font=\normalsize, rounded corners=5pt, line width=0.9pt,
      drop shadow={shadow xshift=0.6mm, shadow yshift=-0.6mm, fill=black!12}
    },
    param/.style={
      rectangle, draw=gray!35, fill=gray!4,
      minimum height=0.55cm, align=center,
      font=\small, rounded corners=8pt, inner xsep=5pt, line width=0.4pt
    },
    bluearr/.style={-{Stealth[length=7pt, width=5pt]}, line width=1.3pt, MidnightBlue!70},
    redarr/.style={-{Stealth[length=7pt, width=5pt]}, line width=1.3pt, BrickRed!80},
    greenarr/.style={-{Stealth[length=7pt, width=5pt]}, line width=1.3pt, OliveGreen!80},
    annot/.style={
      font=\small\itshape, text=MidnightBlue!70,
      fill=white, draw=MidnightBlue!12, rounded corners=2pt,
      inner sep=2.5pt, line width=0.3pt
    },
  ]

    \node[iconnode] (uav) {%
      \includegraphics[width=1.3cm]{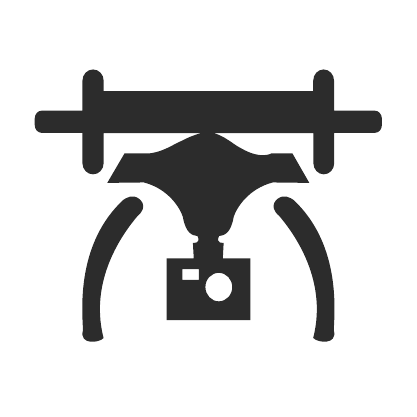}};
    \node[below=0.15cm of uav, font=\normalsize\bfseries, text=MidnightBlue!85] (uavlabel) {UAV Camera};
    \node[below=-0.02cm of uavlabel, font=\small, text=gray!70] {GPS-Denied};

    \node[box, right=1.6cm of uav] (detect) {\textbf{YOLO11\textit{l}}\\[-1pt]\textbf{-OBB}};
    \node[box, right=1.6cm of detect] (outlier) {\textbf{Outlier}\\[-1pt]\textbf{Filtering}};
    \node[box, right=1.6cm of outlier] (kde) {\textbf{KDE Mode}\\[-1pt]\textbf{Estimation}};
    \node[box, right=1.6cm of kde] (gsd) {\textbf{Resolution}\\[-1pt]\textbf{Guard \& GSD}};

    \begin{scope}[on background layer]
      \node[skillbox, fit=(detect)(outlier)(kde)(gsd),
            inner xsep=20pt, inner ysep=28pt,
            label={[font=\normalsize\bfseries, text=MidnightBlue!85, yshift=2pt]above:{Deterministic Geometric Skill (API)}}] (skill) {};
    \end{scope}

    \node[param, below=0.35cm of detect] {Vehicle Det.};
    \node[param, below=0.35cm of outlier] {$\alpha = 1.5$};
    \node[param, below=0.35cm of kde] {$\Lref = 5.045$\,m};
    \node[param, below=0.35cm of gsd] {$\Pthresh \approx 17$\,px};

    \draw[bluearr] (uav) -- node[annot, above, pos=0.5] {image} (detect);
    \draw[bluearr] (detect) -- node[annot, above] {$\{P_i, C_i\}$} (outlier);
    \draw[bluearr] (outlier) -- node[annot, above] {filtered} (kde);
    \draw[bluearr] (kde) -- node[annot, above] {$\Pmode$} (gsd);

    \node[iconnode=violet, right=2.4cm of gsd] (brain) {%
      \includegraphics[width=1.3cm]{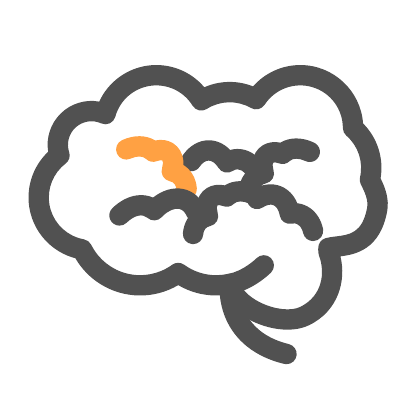}};
    \node[below=0.15cm of brain, font=\normalsize\bfseries, text=violet!80] (brainlabel) {LLM / VLM};
    \node[below=-0.02cm of brainlabel, font=\small, text=gray!70] {Planner};

    \draw[greenarr] (gsd) -- node[annot, above, text=OliveGreen!70] {$\GSDpred$, $\mathcal{C}$} (brain);

    \node[outcome=BrickRed, above=2.8cm of outlier] (vguess) {\textbf{VLM Direct}\\[-1pt]\textbf{Visual Guess}};
    \node[outcome=BrickRed, right=2.0cm of vguess] (halluc) {\textbf{Spatial Scale}\\[-1pt]\textbf{Hallucination}};

    \draw[redarr] (uav.north) |- node[pos=0.78, above, font=\small\itshape, text=BrickRed!70] {raw pixels only} (vguess.west);
    \draw[redarr] (vguess) -- node[annot, above, text=BrickRed!70] {$>$50\% error} (halluc);
    \draw[redarr] (halluc.east) -| (brain.north);

    \node[font=\LARGE\bfseries, text=BrickRed] at ($(halluc.south east)+(0.35, -0.15)$) {\texttimes};

    \node[outcome=OliveGreen, right=1.8cm of brain] (safe) {\textbf{Safe Metric}\\[-1pt]\textbf{Planning}};
    \draw[greenarr] (brain) -- node[annot, above, text=OliveGreen!70] {grounded} (safe);

    \node[font=\LARGE\bfseries, text=OliveGreen] at ($(safe.north east)+(0.35, 0.15)$) {\checkmark};

  \end{tikzpicture}%
  }
  \caption{System architecture as an Embodied Agent Perception Loop. \textbf{Left}: a UAV operating in a GPS-denied environment captures monocular imagery. \textbf{Centre} (dashed box): our Deterministic Geometric Skill processes the image through YOLO-OBB detection, outlier filtering, KDE mode estimation, and a resolution guard to produce a calibrated GSD with confidence score. \textbf{Right}: the LLM/VLM Planner can either rely on direct visual estimation (\textcolor{BrickRed}{red path}), suffering from Spatial Scale Hallucination ($>$50\% error), or invoke our Geometric Skill (\textcolor{OliveGreen}{green path}) for safe metric planning.}
  \label{fig:pipeline}
\end{figure*}

Given a monocular aerial image $I$ with unknown GSD, our Geometric Perception Skill estimates $\GSDpred$ (metres per pixel) using only the image content.
The skill is designed as a \emph{stateless, deterministic API}: a calling agent submits an image and receives a structured response $(\GSDpred, \mathcal{C})$, where $\mathcal{C} \in [0,1]$ is a composite confidence score that the agent can use to gate downstream decisions.
The pipeline consists of five stages (Fig.~\ref{fig:pipeline}):
(1)~vehicle detection via OBB,
(2)~outlier filtering,
(3)~KDE mode estimation,
(4)~GSD computation, and
(5)~multi-dimensional confidence evaluation.

\subsection{Reference Length Determination}
\label{sec:lref}

The reference vehicle length $\Lref$ is a critical parameter.
Rather than selecting an arbitrary value, we derive $\Lref$ statistically from the DOTA~v1.5 training set, which contains 125{,}977 annotated ``small-vehicle'' instances with known GSD.
For each annotation, we compute the physical vehicle length as:
\begin{equation}
  L_i = P_i^{\mathrm{(gt)}} \times \mathrm{GSD}_i^{\mathrm{(gt)}}
\end{equation}
where $P_i^{\mathrm{(gt)}}$ is the OBB longer-side pixel length and $\mathrm{GSD}_i^{\mathrm{(gt)}}$ is the ground-truth GSD of the corresponding image.
We apply KDE to the distribution of $\{L_i\}$ and identify the mode:
\begin{equation}
  \Lref = \arg\max_x \sum_{i=1}^{N} K_h(x - L_i) = 5.045 \text{ m}
\end{equation}
This value represents the most common vehicle length in the dataset, corresponding to typical sedan-class vehicles.
We note that $\Lref$ is dataset-specific: the DOTA training set primarily covers Chinese urban scenes, and vehicle fleets in other regions may exhibit different modal lengths (see Limitations in Section~\ref{sec:discussion}).
The sensitivity of GSD estimation to $\Lref$ is analysed in Section~\ref{sec:ablation}.

\subsection{Vehicle Detection}
\label{sec:detection}

We employ oriented bounding box (OBB) detection to identify small vehicles in the input image.
Unlike axis-aligned detection, OBB tightly encloses each vehicle regardless of its orientation, providing accurate pixel-length measurements.

The detector outputs a set of $N$ detections $\{(P_i, C_i)\}_{i=1}^N$, where $P_i$ is the longer side of the OBB in pixels and $C_i \in [0, 1]$ is the detection confidence score.
Detections with $C_i < 0.1$ are discarded.

For large remote sensing images (e.g., $4000 \times 4000$ pixels), we apply a tiling strategy: the image is divided into overlapping $1024 \times 1024$ tiles with 25\% overlap (stride 768~px), detection is run on each tile, and results are merged with oriented-box non-maximum suppression (IoU threshold 0.45) to handle tile-boundary duplicates.

\subsection{Outlier Filtering}
\label{sec:outlier}

Vehicle detections may include false positives (e.g., rooftop structures, road markings) or genuine vehicles of atypical size (buses, trucks).
We apply a simple median-based outlier filter:
\begin{equation}
  \text{Keep detection } i \quad \text{iff} \quad P_i \leq \alpha \cdot \mathrm{median}(\{P_j\}_{j=1}^N)
\end{equation}
where $\alpha = 1.5$ is the outlier factor.
This preserves the majority of small vehicles while removing large outliers that would distort the pixel-length distribution.
The sensitivity to $\alpha$ is examined in Section~\ref{sec:ablation}.

\subsection{KDE Mode Estimation}
\label{sec:kde}

Given the filtered pixel lengths $\{P_i\}_{i=1}^{N'}$, we estimate the modal pixel length $\Pmode$ using kernel density estimation.
The density function is:
\begin{equation}
  \hat{f}(x) = \frac{1}{N'} \sum_{i=1}^{N'} K_h(x - P_i)
  \label{eq:kde}
\end{equation}
where $K_h(\cdot)$ is a Gaussian kernel with bandwidth $h$ selected via Scott's rule.
The modal pixel length is:
\begin{equation}
  \Pmode = \arg\max_x \hat{f}(x)
\end{equation}

Unlike simple summary statistics (mean, median), KDE captures the full distributional shape and identifies the true mode, making it robust to skewed distributions caused by false positives or atypical vehicles.
A weighted variant using detection confidence $C_i$ as KDE weights is evaluated in the ablation study (Section~\ref{sec:ablation}) but provides no measurable improvement over unweighted KDE.

When fewer than 5 detections survive filtering ($N' < 5$), KDE is unreliable due to insufficient samples.
In this regime, we fall back to the simple median of pixel lengths.

Finally, GSD is computed as:
\begin{equation}
  \GSDpred = \frac{\Lref}{\Pmode}
  \label{eq:gsd}
\end{equation}

\subsection{Confidence Evaluation and Safety Fallback}
\label{sec:confidence}

Each GSD prediction is accompanied by a composite confidence score $\mathcal{C} \in [0, 1]$ aggregating four dimensions via a weighted sum: (1)~sample sufficiency ($w = 0.35$), penalising few detections; (2)~distribution concentration ($w = 0.35$), rewarding low coefficient of variation; (3)~detection quality ($w = 0.20$), reflecting median YOLO confidence; and (4)~anomaly detection ($w = 0.10$), flagging physically implausible pixel lengths.

\paragraph{Autonomous safety fallback.}
The composite score is blind to a critical physical limitation: at coarse resolution (GSD~$\gtrsim 0.3$\,m/px), vehicles are detected with consistent but systematically compressed pixel lengths, causing the score to remain high despite large GSD errors.
For safe autonomous operation, we augment $\mathcal{C}$ with a hard resolution guard:
\begin{equation}
  \mathcal{C} = \begin{cases}
    \mathcal{C}_{\mathrm{raw}} & \text{if } \Pmode \geq \Pthresh \\
    \min(\mathcal{C}_{\mathrm{raw}},\; 0.3) & \text{if } \Pmode < \Pthresh
  \end{cases}
  \label{eq:resolution_guard}
\end{equation}
where $\Pthresh = \Lref / \GSDmax$ with $\GSDmax = 0.3$\,m/px, yielding $\Pthresh \approx 17$\,pixels.
This threshold adapts automatically when $\Lref$ is recalibrated for a different region.
A UAV planner receiving $\mathcal{C} < 0.5$ should autonomously fall back to alternative localisation strategies (e.g., altitude hold, visual odometry) rather than trusting the GSD estimate---this confidence-gated design allows the calling agent to maintain a safe decision loop without human intervention.
Concretely, the skill returns a structured record $\{\GSDpred, \mathcal{C}, N', \Pmode\}$ rather than a free-form answer; the special case $\mathcal{C} = 0$ (no detectable anchor, 33\% of DOTA images) signals that \emph{no} metric estimate is available, so the planner withholds scale-dependent decisions and queries an alternative anchor instead of acting on a hallucinated guess---this is what we mean by ``safe metric planning.''

\section{EXPERIMENTS}
\label{sec:experiments}


\begin{table*}[!t]
  \centering
  \caption{GSD estimation results on DOTA~v1.5 validation set (458~images total). The GT Baseline uses ground-truth annotations (269~images). The ``intersection'' rows restrict both E2E and GT to the 267~images covered by both evaluations, enabling a like-for-like comparison.}
  \label{tab:e2e}
  \begin{tabular}{lccccc}
    \toprule
    \textbf{Detector} & \textbf{Evaluated} & \textbf{Median Err} & \textbf{Mean Err} & $\boldsymbol{<}$\textbf{10\%} & $\boldsymbol{<}$\textbf{20\%} \\
    \midrule
    YOLOv8m-OBB (v1) & 310 & 7.17\% & 14.88\% & 63.2\% & 80.0\% \\
    \YOLOviiil{} (v2)  & 310 & 6.87\% & 13.44\% & 65.2\% & 82.3\% \\
    \YOLOxil{} (v3)  & 306 & 6.87\% & 12.89\% & 66.0\% & 83.3\% \\
    GT Baseline (all)              & 269 & 6.88\% & 8.54\% & 67.7\% & 92.6\% \\
    \midrule
    \YOLOxil{} (v3, intersection) & 267 & \textbf{6.22\%} & \textbf{9.41\%} & \textbf{71.9\%} & \textbf{89.5\%} \\
    GT Baseline (intersection)     & 267 & 6.88\% & 8.56\% & 67.4\% & 92.5\% \\
    \bottomrule
  \end{tabular}
\end{table*}

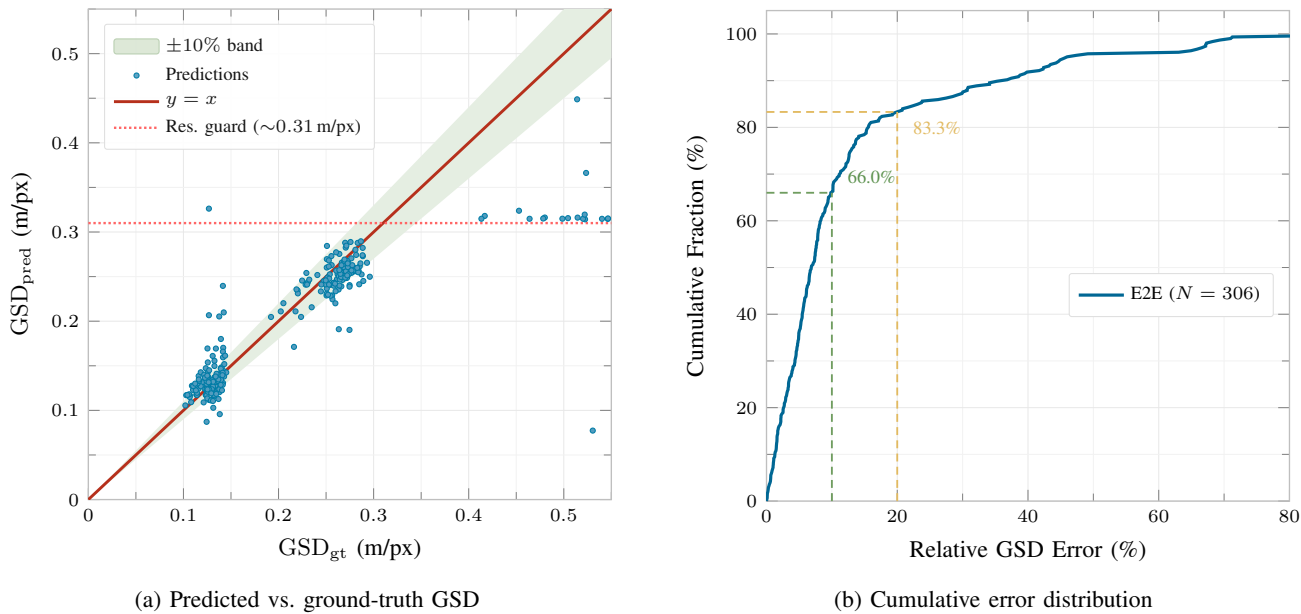
\begin{figure*}[!t]
  \centering
  \begin{minipage}[b]{0.48\textwidth}
    \centering
    \begin{tikzpicture}
    \begin{axis}[
        width=\textwidth,
        height=0.95\textwidth,
        xlabel={$\GSDgt$ (m/px)},
        ylabel={$\GSDpred$ (m/px)},
        xlabel style={font=\small},
        ylabel style={font=\small},
        xmin=0, xmax=0.55,
        ymin=0, ymax=0.55,
        grid=both,
        minor tick num=1,
        grid style={gray!10, line width=0.15pt},
        major grid style={gray!18, line width=0.35pt},
        tick label style={font=\scriptsize},
        axis line style={gray!50, line width=0.5pt},
        mark size=1.2pt,
        legend pos=north west,
        legend style={
          font=\scriptsize, fill=white, fill opacity=0.92,
          draw=gray!25, rounded corners=1pt,
          inner sep=3pt, row sep=0.5pt,
          /tikz/every even column/.append style={column sep=4pt}
        },
        legend cell align={left},
    ]
    \addplot[name path=upper, draw=none, forget plot, domain=0:0.55, samples=2] {1.1*x};
    \addplot[name path=lower, draw=none, forget plot, domain=0:0.55, samples=2] {0.9*x};
    \addplot[fill=OliveGreen!10, draw=none, forget plot] fill between[of=upper and lower];
    \addlegendimage{area legend, fill=OliveGreen!15, draw=OliveGreen!30}
    \addlegendentry{$\pm 10\%$ band}

    \addplot[only marks, mark=*, mark options={fill=MidnightBlue!55, draw=MidnightBlue!80, scale=0.8}]
        table[x=gsd_gt, y=gsd_pred] {gsd_scatter.dat};

    \addplot[domain=0:0.55, BrickRed, line width=1.1pt, no markers] {x};

    \addplot[domain=0:0.55, red!60, densely dotted, line width=0.9pt, no markers] {0.31};

    \legend{$\pm 10\%$ band, Predictions, $y = x$, Res.\ guard (${\sim}0.31$\,m/px)}
    \end{axis}
    \end{tikzpicture}
    \\[2pt] {\small (a) Predicted vs.\ ground-truth GSD}
  \end{minipage}
  \hfill
  \begin{minipage}[b]{0.48\textwidth}
    \centering
    \begin{tikzpicture}
    \begin{axis}[
        width=\textwidth,
        height=0.95\textwidth,
        xlabel={Relative GSD Error (\%)},
        ylabel={Cumulative Fraction (\%)},
        xlabel style={font=\small},
        ylabel style={font=\small},
        tick label style={font=\scriptsize},
        axis line style={gray!50, line width=0.5pt},
        xmin=0, xmax=80,
        ymin=0, ymax=105,
        grid=both,
        minor tick num=1,
        grid style={gray!10, line width=0.15pt},
        major grid style={gray!18, line width=0.35pt},
        legend style={
          font=\scriptsize, fill=white, draw=gray!25,
          rounded corners=1pt, at={(0.97,0.42)}, anchor=east,
          inner sep=3pt, row sep=0.5pt
        },
        legend cell align={left},
    ]
    \addplot[MidnightBlue, line width=1.2pt, no markers]
        table[x=error, y=cumfrac] {gsd_cdf.dat};

    \addplot[OliveGreen!70, densely dashed, line width=0.7pt, forget plot]
        coordinates {(10, 0) (10, 66)};
    \addplot[OliveGreen!70, densely dashed, line width=0.7pt, forget plot]
        coordinates {(0, 66) (10, 66)};
    \addplot[Goldenrod!50!brown, densely dashed, line width=0.7pt, forget plot]
        coordinates {(20, 0) (20, 83.3)};
    \addplot[Goldenrod!50!brown, densely dashed, line width=0.7pt, forget plot]
        coordinates {(0, 83.3) (20, 83.3)};

    \node[font=\scriptsize, text=OliveGreen!80, anchor=south west] at (axis cs:11, 66) {66.0\%};
    \node[font=\scriptsize, text=Goldenrod!50!brown, anchor=north west] at (axis cs:21, 83.3) {83.3\%};

    \legend{E2E ($N = 306$)}
    \end{axis}
    \end{tikzpicture}
    \\[2pt] {\small (b) Cumulative error distribution}
  \end{minipage}
  \caption{End-to-end GSD estimation results (\YOLOxil{}, 306~images). (a)~Predicted vs.\ ground-truth GSD for images with GSD~$< 0.55$\,m/px (271 of 306 images). The shaded band denotes $\pm 10\%$ error; the horizontal dotted line at $\GSDpred \approx 0.31$\,m/px marks the resolution guard limit ($\Pmode \approx 16$\,px), beyond which predictions saturate. (b)~Cumulative error distribution: 66.0\% of images achieve $< 10\%$ error (green dashed) and 83.3\% achieve $< 20\%$ error (amber dashed). The long tail corresponds primarily to low-resolution images (GSD~$> 0.3$\,m/px).}
  \label{fig:e2e_results}
\end{figure*}

\begin{table*}[!t]
\centering
\caption{Area estimation on RS-GSD Benchmark v5.0 (100~entries, 8~categories). Upper: zero-shot VLMs. Middle: VLMs with vehicle-length hint. Lower: our pipeline.}
\label{tab:area_comparison}
\begin{tabular}{lcccccc}
\toprule
\textbf{Model} & \textbf{Med.\ Err.} & \textbf{Mean Err.} & \textbf{$<$10\%} & \textbf{$<$25\%} & \textbf{$<$50\%} & \textbf{$<$100\%} \\
\midrule
\multicolumn{7}{l}{\emph{Zero-shot (no scale information):}} \\
Qwen2.5-VL-72B            & 38.3\% & 84.5\%  & 19\% & 35\% & 61\% & 84\% \\
Claude Opus 4.6            & 47.2\% & 72.9\%  & 24\% & 35\% & 55\% & 73\% \\
Qwen-VL-Max               & 50.3\% & 97.1\%  & 13\% & 27\% & 49\% & 75\% \\
Qwen-VL-Plus              & 51.3\% & 190.3\% & 22\% & 33\% & 49\% & 68\% \\
\GPTfo{}                   & 51.9\% & 69.4\%  & 17\% & 34\% & 48\% & 93\% \\
\midrule
\multicolumn{7}{l}{\emph{With vehicle-length hint (``cars $\approx$ 5\,m''):}} \\
Claude Opus 4.6 + hint$^\dagger$ & 17.1\% & 41.5\%  & 36\% & 59\% & 73\% & 88\% \\
\GPTfo{} + hint            & 35.9\% & 55.2\%  & 15\% & 36\% & 59\% & 85\% \\
Qwen2.5-VL-72B + hint     & 38.1\% & 100.4\% & 10\% & 30\% & 59\% & 83\% \\
Qwen-VL-Max + hint        & 46.5\% & 99.9\%  & 13\% & 30\% & 53\% & 80\% \\
Qwen-VL-Plus + hint       & 58.0\% & 236.2\% & 12\% & 28\% & 44\% & 70\% \\
\midrule
RS-GSD (ours)             & \textbf{19.7\%} & \textbf{29.2\%}  & \textbf{30\%} & \textbf{59\%} & \textbf{80\%} & \textbf{97\%} \\
\bottomrule
\end{tabular}
\vspace{1mm}
\par\noindent{\footnotesize $^\dagger$Claude Opus~4.6 operates as an \emph{agent} with programmatic image analysis tools, unlike other VLMs which perform pure visual estimation.}
\end{table*}

\begin{figure*}[!t]
  \centering
  \includegraphics[width=\textwidth]{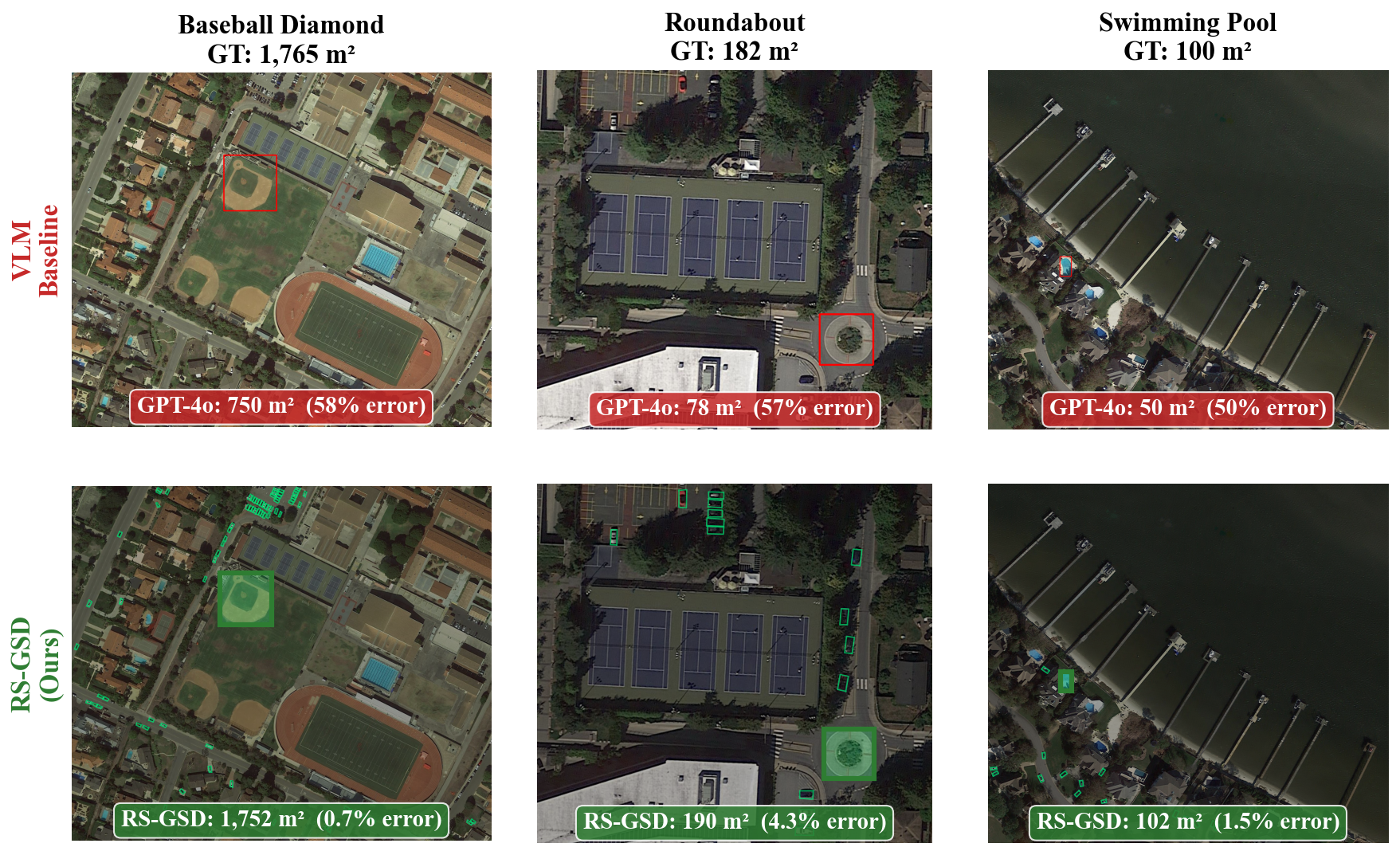}
  \caption{Qualitative comparison on three benchmark entries. \textbf{Top}: annotated input with \GPTfo{} predictions (red). \textbf{Bottom}: RS-GSD pipeline with vehicle OBBs (green) and SAM segmentation overlay. Our geometric approach achieves 0.7--4.3\% error vs.\ \GPTfo{}'s 50--58\% error, illustrating the spatial scale hallucination phenomenon.}
  \label{fig:qualitative}
\end{figure*}

\subsection{Dataset and Setup}
\label{sec:dataset}

We evaluate on the DOTA~v1.5 dataset \cite{xia2018dota, ding2024dotav15}. Although primarily comprising high-resolution satellite imagery, DOTA serves as a standard, rigorous proxy for high-altitude UAV nadir views due to its equivalent perspective and precise GSD metadata. The dataset contains high-resolution aerial images with oriented bounding box annotations for 16~object categories.
We use the ``small-vehicle'' category for training the vehicle detector.
Critically, DOTA images include GSD metadata, enabling quantitative evaluation of our GSD predictions.

The dataset comprises 1{,}411 training images and 458 validation images.
For detector training, we split the training set into an 85\% Training Subset and a 15\% Dev Set for hyperparameter tuning.
The full 458-image validation set serves as our test set.
Of these, 8~images lack GSD metadata and are excluded, leaving 450~evaluable images.

We evaluate GSD estimation accuracy using the relative error:
\begin{equation}
  \epsilon_{\mathrm{rel}} = \frac{|\GSDpred - \GSDgt|}{\GSDgt}
  \label{eq:error}
\end{equation}

We report median and mean relative error, as well as the proportion of images with $\epsilon_{\mathrm{rel}} < 10\%$ and $\epsilon_{\mathrm{rel}} < 20\%$.

\subsection{Vehicle Detection and GSD Evaluation}
\label{sec:yolo_training}

We train \YOLOxil{}~\cite{yolo11_ultralytics} on the DOTA~v1.5 Training Subset for oriented vehicle detection, achieving mAP50 = 0.660 on the Dev Set.
Using ground-truth annotations as an upper bound, the statistical pipeline achieves 6.88\% median GSD error on 269~images---the irreducible error from vehicle length variability (Table~\ref{tab:e2e}).

End-to-end, replacing GT annotations with YOLO detections, the pipeline achieves 6.87\% median error on 306~images (67\% coverage).
On the 267-image intersection of E2E and GT evaluation sets, E2E achieves 6.22\% median error vs.\ the GT baseline's 6.88\%, confirming that KDE mode estimation effectively absorbs detection noise.
Figure~\ref{fig:e2e_results}(a)~plots predicted vs.\ ground-truth GSD, and Fig.~\ref{fig:e2e_results}(b)~the cumulative error distribution.
The pipeline produces no estimate for 152~images (33\%) lacking detectable vehicles; these receive $\mathcal{C} = 0$.

The gap between E2E and GT is larger for mean error (12.89\% vs.\ 8.54\%) due to images with $<$5 detections where the fallback median estimator is less accurate.

\subsection{Ablation and Sensitivity Analysis}
\label{sec:ablation}

Table~\ref{tab:ablation} consolidates the key ablation findings.

\begin{table}[t]
  \centering
  \caption{Ablation and sensitivity analysis. All results use GT annotations (269~images) unless noted. ``E2E'' uses \YOLOxil{} detections (306~images).}
  \label{tab:ablation}
  \begin{tabular}{llcc}
    \toprule
    \textbf{Factor} & \textbf{Setting} & \textbf{Med.\ Err} & \textbf{$<$20\%} \\
    \midrule
    \multicolumn{4}{l}{\emph{Aggregation method (E2E):}} \\
    & KDE (ours)     & \textbf{6.87\%} & 83.3\% \\
    & Median         & 7.31\% & 84.0\% \\
    & Mean           & 8.27\% & 83.7\% \\
    \midrule
    \multicolumn{4}{l}{\emph{Reference length $\Lref$:}} \\
    & 4.5\,m         & 10.59\% & 80.3\% \\
    & 5.045\,m (ours) & \textbf{6.88\%} & 92.6\% \\
    & 5.5\,m         & 10.71\% & 79.9\% \\
    \midrule
    \multicolumn{4}{l}{\emph{Outlier factor $\alpha$:}} \\
    & 1.0            & 7.27\% & 87.7\% \\
    & 1.5 (ours)     & \textbf{6.88\%} & 92.6\% \\
    & None           & 6.74\% & 92.6\% \\
    \midrule
    \multicolumn{4}{l}{\emph{Vehicle count (E2E):}} \\
    & $N \geq 20$    & 6.12\% & --- \\
    & $N < 5$        & 13.47\% & --- \\
    \midrule
    \multicolumn{4}{l}{\emph{GSD range (E2E):}} \\
    & $< 0.3$\,m/px  & \textbf{6.10\%} & --- \\
    & $> 0.7$\,m/px  & 67.19\% & --- \\
    \bottomrule
  \end{tabular}
\end{table}

\paragraph{Key findings.}
KDE mode estimation outperforms mean aggregation by 17\% relative improvement (6.87\% vs.\ 8.27\% E2E median error), as KDE captures the distributional shape and is robust to outliers.
The reference length $\Lref$ is the most sensitive parameter: a $\pm$0.5\,m deviation roughly doubles the median error, validating data-driven calibration.
Performance scales with vehicle count ($\geq$20 vehicles: 6.12\%; $<$5: 13.47\%) and degrades sharply at coarse resolution ($>$0.7\,m/px: 67\%), confirming the method's operating envelope of sub-metre urban imagery.
The confidence score ($r = -0.354$ with error) provides coarse quality gating; the resolution guard correctly flags 35 of 51 high-error images.
Among the four confidence terms, this resolution guard is the most consequential safety trigger, whereas the sample-sufficiency term mainly governs the sparse-detection regime ($N' < 5$) that is handed to the median fallback.

\subsection{Spatial Scale Hallucination: RS-GSD vs.\ VLMs}
\label{sec:area_measurement}

We first fix terminology: \textbf{VehAnchor} denotes the deterministic GSD-estimation skill of Section~\ref{sec:method}, while \emph{RS-GSD} (Remote Sensing GSD) names both the evaluation benchmark introduced below and the full area-measurement pipeline (VehAnchor\,+\,SAM) whose results are labelled ``RS-GSD~(Ours)'' in Table~\ref{tab:area_comparison} and Fig.~\ref{fig:qualitative}.

To quantify VLM spatial hallucinations, we benchmark area estimation: $A = N_{\mathrm{px}} \times \mathrm{GSD}^2$.
The \emph{RS-GSD Benchmark v5.0} (100~entries, 64~images, 8~categories from iSAID~\cite{waqas2019isaid} $\cap$ DOTA~v1.5) deliberately oversamples irregular objects (swimming pools, roundabouts) to stress-test memorised-size priors.
We evaluate five VLMs~\cite{bai2025qwen25vl, anthropic2024claude, wang2024qwen2vl, openai2024gpt4o} in zero-shot and vehicle-length-hinted settings\footnote{All VLMs receive an identical fixed query: a system role (``a remote sensing image analysis expert'') and a user message asking for the area, in $\mathrm{m}^2$, of the object inside a red rectangle, ending with a machine-parseable \texttt{AREA=$\langle$number$\rangle$} line. The hinted setting appends one sentence: ``small vehicles (cars) are approximately 5\,m long; use them as a scale reference.''}; our pipeline uses YOLO-OBB + SAM~\cite{kirillov2023sam} segmentation.

\paragraph{Results.}
Table~\ref{tab:area_comparison} reveals severe spatial hallucinations: all zero-shot VLMs cluster at 38--52\% median error.
With a vehicle-length hint (``cars $\approx$ 5\,m''), only Claude (17.1\%, using programmatic tools$^\dagger$) and \GPTfo{} (35.9\%) improve meaningfully; Qwen models show negligible or negative change.
Our RS-GSD pipeline achieves 19.7\% median error with 2.6$\times$ lower category dependence (polarisation ratio 1.9$\times$ vs.\ 5.0$\times$) and 4$\times$ fewer catastrophic failures ($<$100\%: 97\% vs.\ 88\%).
The tight median-mean gap of our pipeline (19.7\% vs.\ 29.2\%) versus VLMs (e.g., 38.3\% vs.\ 84.5\% for Qwen2.5-VL-72B) confirms that geometric measurement is fundamentally more reliable than visual estimation.

\paragraph{Computation and agency.}
All ``RS-GSD~(Ours)'' areas are computed geometrically as $A = N_{\mathrm{px}} \times \GSDpred^2$ from the SAM mask---never queried from a VLM---so accuracy depends jointly on the estimated GSD and the segmentation; the low error on irregular shapes (Fig.~\ref{fig:qualitative}) indicates SAM masks are not the dominant error term.
Because this pipeline applies the geometric tool unconditionally and reports its output without any LLM spatial reasoning, the ``RS-GSD~(Ours)'' row also answers what an agent obtains by always invoking the tool and trusting it at face value; the confidence score then re-introduces agency, letting a deployed agent decline the estimate when $\mathcal{C} < 0.5$.

\section{DISCUSSION}
\label{sec:discussion}

\paragraph{Deterministic vs.\ learned spatial reasoning.}
Our experiments demonstrate a fundamental gap between VLM-based spatial reasoning and deterministic geometric measurement.
KDE mode estimation outperforms mean aggregation by 17\%, and even with explicit vehicle-length hints, most VLMs cannot translate this prior into accurate area estimates---confirming the spatial scale hallucination phenomenon~\cite{liu2023vsr, chen2024spatialvlm, liao2024qspatial}.
This finding supports the emerging \emph{tool-augmented agent} paradigm: rather than expecting end-to-end models to master every perceptual modality, we equip LLM/VLM planners with specialised, deterministic tools for tasks that demand metric precision.
VehAnchor instantiates this paradigm for spatial scale: the planner delegates GSD estimation to our geometric skill and receives a confidence-gated result, preserving the agent's high-level reasoning capability while grounding its spatial understanding in physics.

\paragraph{Limitations.}
The method requires small vehicles in the scene (absent in 33\% of DOTA images) and sub-metre resolution (GSD~$\lesssim 0.3$\,m/px).
The reference length $\Lref = 5.045$\,m is calibrated for Chinese urban fleets; deployment in other regions requires recalibration. Our $\Lref \in \{4.5, 5.045, 5.5\}$\,m sweep (Table~\ref{tab:ablation}) brackets typical regional fleets---e.g., larger North-American SUVs/pickups ($\sim$5.5\,m) versus smaller European subcompacts ($\sim$4.5\,m)---and shows a $\pm$0.5\,m mismatch roughly doubles median error, quantifying the calibration overhead for a new region.
Oblique or off-nadir imagery introduces perspective distortion not modelled by our pipeline: a camera pitch $\theta$ from nadir foreshortens along-track pixel length by $\approx\cos\theta$, biasing GSD by the same factor (e.g., ${\sim}6\%$ at $20^\circ$), so the method targets near-nadir views.
Extending to multiple reference object classes (e.g., road lane widths, shipping containers) could improve coverage.
Geospatial foundation models~\cite{liu2023remoteclip, kuckreja2024geochat} offer an anchor-free alternative, but they inherit the same metric-grounding gap as general VLMs and would still require explicit scale calibration; benchmarking them as a baseline is left to future work.

\section{CONCLUSION}
\label{sec:conclusion}

We have presented VehAnchor, a lightweight, deterministic Geometric Perception Skill that equips tool-augmented embodied agents with the ability to recover absolute metric scale from monocular aerial imagery without GPS or camera metadata.
By detecting ubiquitous small vehicles and estimating their modal pixel length via KDE, our method achieves 6.87\% median GSD error on DOTA~v1.5 (306~images).
Integrated with SAM segmentation for area measurement, the pipeline yields 19.7\% median error on a 100-entry benchmark---with 2.6$\times$ lower category dependence and 97\% of predictions within 100\% error---while five state-of-the-art VLMs exhibit severe spatial scale hallucinations (38--52\% median error zero-shot).
These results demonstrate that the tool-augmented agent paradigm---where LLM planners delegate metric-critical perception to deterministic geometric skills---is essential for safe autonomous spatial reasoning.
Future work includes extending to multiple reference object classes, validating on diverse geographic regions, and integrating VehAnchor into closed-loop UAV planning systems as a standard perception tool alongside depth estimation and obstacle detection.

\addtolength{\textheight}{-1cm}   
\bibliographystyle{IEEEtran}
\bibliography{references}

\end{document}